\documentclass[11pt]{article}

\usepackage[preprint]{acl}

\usepackage{times}
\usepackage{latexsym}
\usepackage[T1]{fontenc}
\usepackage[utf8]{inputenc}
\usepackage{microtype}
\usepackage{inconsolata}
\usepackage{graphicx}
\usepackage{fontawesome5}  
\usepackage{xurl}          

\usepackage{tikz}
\usetikzlibrary{arrows.meta,positioning,fit,backgrounds,calc}

\title{\faCoins~Paying to Know: Micro-Transaction Markets for Verified Product\\
Information in Agentic E-Commerce}

\author{Filippos Ventirozos, Matthew Shardlow \\
  Manchester Metropolitan University \\
  Manchester, United Kingdom \\
  \texttt{f.ventirozos@mmu.ac.uk}}

\begin{document}
\maketitle

\begin{abstract}
Commercial NLP treats the shopping chatbot as a recommender or a conversion tool: its job is to match a user to a catalogue entry and close a sale. We argue that the arrival of agent-native micro-payment rails (e.g., x402, AP2) changes what is scarce. When the buyer is an autonomous agent that can investigate exhaustively, the bottleneck is no longer \emph{matching} products but \emph{acquiring trustworthy, decision-relevant information} about them. We envision agentic e-commerce as a micro-transaction market for verified information: buyer agents spend fractions of a cent to progressively unlock seller- and reviewer-supplied data---service histories, third-party test reports, bills of materials, audited sales and support metrics---paid for \`a la carte under a freemium model, with reviewer trust scored reputationally. We sketch the architecture of such a market and argue that it rewards genuine product quality and yields truer competition than ranking-based storefronts. We then translate the vision into concrete NLP problems---cost-optimal information acquisition, data pricing and negotiation, real-time entity resolution, grounded value exchange, and privacy-preserving persona modelling---and argue that these, not chat fluency, deserve the field's attention.


\end{abstract}



\section{Introduction}

Conversational agents are moving from \emph{answering} product questions to \emph{acting} on them. With agent-native payment standards such as Google's Agent Payments Protocol \citep{google2025ap2} and emerging agent-to-agent economic infrastructure utilizing x402 micropayments \citep{vaziry2025a2a}, an LLM can now hold a budget and settle a purchase on its owner's behalf---at sub-cent granularity, without human intervention. The storefront is becoming an API, and the buyer is becoming a program.

Yet applied NLP still largely studies these systems as recommenders and conversion engines---agents that match a shopper to a catalogue entry and persuade them to buy \citep{yu2026shopping,salvi2026persuasion}. This framing quietly assumes that the hard part is matching. We contend that once buyers are tireless agents that can query, compare, and verify at scale, matching becomes cheap; what stays scarce is \emph{trustworthy, decision-relevant information} about a product. Today that information is either free but unverified (SEO-optimised copy, unaudited star ratings) or locked in silos, and discovery is mediated by a marketplace whose ranking incentives need not align with the buyer's.

\paragraph{Position.} We take the position that \emph{in agentic e-commerce, value migrates from matching products to pricing and trading verified information about them}. We envision a \emph{micro-transaction information market} in which a buyer's agent spends micro-payments to progressively unlock data that a seller (or a third-party reviewer) has chosen to expose, paid \`a la carte under a freemium model. A product that withstands scrutiny can afford to reveal more; an agent that prices its data well earns proportionally more. The result, we argue, is a market that rewards genuine quality and approximates true competition, rather than ``the first sponsored result.''

\paragraph{Two running examples.} The two scenarios below are illustrative rather than a survey: both are automotive, chosen to span a high- and a low-stakes purchase, and the same mechanism carries to other categories. Both are deliberately idealised; we bracket the ways such a market could be \emph{skewed} and treat the most pressing in our Limitations and Ethical Considerations.\footnote{Among them: collusion among sellers, reviewers, or data vendors; reputation gaming and Sybil-staked reviews; strategic non-disclosure of unfavourable evidence; persona-based price discrimination; and access barriers that entrench incumbents---each beyond the scope of this position paper.}

\faCar~\emph{High stakes: a used car.}
Consider a buyer agent instructed to find a reliable used hatchback under
\pounds9{,}000. The public listing exposes only cheap-to-broadcast facts:
make, model, mileage, location, photographs, and asking price. The buyer agent
then buys evidence from distinct stakeholders, in a progressively costly
trail.\footnote{Prices here are illustrative; what matters is the rising cost of
evidence, not the exact tariff.} From the seller it pays \pounds0.25 for a
redacted service-history digest (dates, mileage, garage identities) and
\pounds1.50 for a richer pack with invoice hashes, diagnostic codes, and
tyre/brake measurements; from registry, finance, or insurer vendors, a small
fee for write-off, stolen, outstanding-finance, and mileage-anomaly checks.
Only if the car remains a candidate does it commit a larger amount---say
\pounds20 as a deposit toward an independent condition report.

Each payment benefits a different party. The buyer reduces hidden risk and
avoids wasted viewings; the seller monetises verified transparency; data
vendors are paid for facts rather than advertising; and the marketplace obtains
structured, comparable evidence. More importantly, the sequence of paid
requests is a better serious-buyer signal than clicks, dwell time, chatbot
engagement, or a refundable viewing reservation---happening currently in marketplaces. A seller can distinguish a
browser from an agent that has already spent \pounds3--\pounds5 resolving the
specific uncertainties that block purchase, and can allocate scarce salesperson
time, document retrieval, test-drive slots, and negotiation effort accordingly.
If a hard constraint is violated---for example an undisclosed write-off marker
or outstanding finance---the buyer agent walks away immediately; if not, its
paid information trail reveals both seriousness and remaining uncertainty.

\faWrench~\emph{Low stakes: an auxiliary belt.} Now consider a snapped auxiliary
belt for a vehicle that the user needs back on the road. The part may cost only
\pounds15--\pounds35, so the agent should not buy expensive reports. Instead, it
caps investigation at a few pence and purchases only small, high-value facts:
fractions of a penny for an OEM part-number cross-reference and VIN/engine-code
fitment confirmation, \pounds0.03 for a certified supplier-reputation profile
(verified-purchase reviews, false-fitment return and dispute rates), and
\pounds0.05 for a materials-and-provenance certificate covering belt compound,
batch identity, and an OEM/licensed-aftermarket authorisation claim.

Here the reputable supplier benefits most. A seller with accurate fitment data,
low return rates, genuine materials, and authorised designs can make those facts
machine-readable and priceable, rather than competing only through SEO,
sponsored placement, or inflated star ratings. The buyer avoids the wrong part;
the seller reduces returns and support queries; the fitment database earns for
a small but decisive datum; and the original manufacturer, designer, or material
supplier can be credited when their contribution genuinely matters. This echoes
the product-passport and responsible-sourcing vision in blockchain provenance
projects, where claims about source, certification, and custody travel with the
physical product rather than being rewritten as marketing copy
\citep{provenance2015blockchain,minespider2023cordis}.

Both examples are, in spirit, a commercial realisation of the Semantic Web:
machine-readable, linked facts about things, but now with provenance, prices,
and negotiation \citep{bernerslee2001semantic}. The goal is fairer
representation. Products should compete on addressable evidence that agents can
verify, price, and compare, not merely on ranking position, advertising budget,
or persuasive copy. The NLP decision is therefore not ``rank the catalogue,''
but decide which stakeholder's evidence is worth buying next, ground it to a
shared schema, and interpret the resulting payment trail as an economic signal.

\paragraph{Frontiers for NLP.} We argue these are frontier problems the applied-NLP community must take up to move forward: meaningful \emph{inter-agent communication}---buyer and seller agents trading priced, grounded, negotiated evidence, as described here---will not emerge from better chat alone, but from progress on cost-optimal acquisition, data pricing, real-time grounding, and verifiable value exchange. Solving them turns conversational shopping into \emph{agentic procurement automation}: agents that investigate, negotiate, and settle purchases end-to-end on a principal's behalf.

\paragraph{Roadmap.} Section~\ref{sec:shift} situates the shift; Section~\ref{sec:vision} develops the vision and its architecture (Figure~\ref{fig:arch}); Section~\ref{sec:challenges} translates it into open NLP problems; Section~\ref{sec:conclusion} concludes.

\section{The Emerging Ecosystem: Agentic Commerce and the Information Gap}
\label{sec:shift}

\paragraph{Agent-native payment rails.} The enabling substrate is programmable payment. The x402 protocol revives the dormant HTTP 402 (``Payment Required'') status code so that a server can demand payment for a resource and a client---human or agent---can settle it inline, in stablecoins, at sub-cent amounts and sub-second latency \citep{vaziry2025a2a}. Google's Agent Payments Protocol (AP2) targets the complementary problem of \emph{authorization}---establishing that an agent's purchase faithfully reflects its owner's intent, and assigning accountability for it \citep{google2025ap2}; in our setting, x402-style rails price the individual data request, while AP2-style authorization binds it to the user's budget and constraints. Payments at this granularity make a previously absurd idea practical: charging a fraction of a cent for a single datum.

\paragraph{A different data paradigm.} This reframes the data problem. Classical e-commerce personalisation infers intent from noisy click-stream signals; conversational commerce promised richer, explicit ``zero-party'' signals volunteered in dialogue. In both cases, however, the \emph{platform harvests the buyer's} data. We invert the flow: the priceable asset is the \emph{seller's verified information about the product}, and the buyer agent is the one paying for it. Where a star rating is a cheap, gameable summary, a cryptographically attestable service record, an audited return-rate, or a paid conversation with an identified prior owner is costly to fake and therefore informative.

\paragraph{The shrinking storefront.} As buyers become agents, the front-facing storefront---the search box, the ranked grid, the ``recommended for you'' carousel---loses its privileged position; agents transact against APIs and negotiate over evidence rather than scroll. The competitive question shifts from \emph{who ranks first?} to \emph{who can substantiate their claims most cheaply and credibly?}

\section{The Vision: A Micro-Transaction Market for Verified Information}
\label{sec:vision}

\begin{figure*}[t]
\centering
\begin{tikzpicture}[
  >={Latex[length=2.4mm]},
  node distance=8mm,
  box/.style={rounded corners=2pt,draw,thick,align=center,minimum height=10mm,inner sep=4pt,font=\small},
  actor/.style={box,fill=blue!8},
  infra/.style={box,fill=gray!12},
  supply/.style={box,fill=green!8},
  nlp/.style={rounded corners=2pt,draw,densely dashed,fill=yellow!18,align=center,font=\scriptsize,inner sep=3pt,text width=33mm},
  flow/.style={->,thick},
  back/.style={->,thick,densely dashed}
]
\node[actor] (user) {User};
\node[actor,right=14mm of user,text width=20mm] (buyer) {Buyer\\Agent};
\node[infra,right=16mm of buyer,text width=26mm] (engine) {Marketplace:\\Listing \&\\Pricing Engine};
\node[supply,right=16mm of engine,text width=26mm] (supply) {Sellers\,$\cdot$\,Data\\vendors\,$\cdot$\,Reviewers};

\draw[flow] (user) -- node[above,font=\scriptsize,align=center] {intent, budget,\\constraints} (buyer);
\draw[flow] (buyer) -- node[above,font=\scriptsize,align=center] {x402/AP2\\micro-payments} (engine);
\draw[flow] (engine) -- node[above,font=\scriptsize,align=center] {evidence\\requests} (supply);
\draw[back] (supply.south) -- ++(0,-9mm) -| node[below,pos=0.25,font=\scriptsize] {verified data payloads} (buyer.south);

\node[nlp,below=12mm of buyer.south west,anchor=north west] (nlpb)
  {\textbf{NLP @ buyer:} preference elicitation; cost-optimal acquisition; grounding check};
\node[nlp,below=12mm of engine.south,anchor=north] (nlpe)
  {\textbf{NLP @ market:} real-time entity resolution; data pricing \& negotiation; reviewer trust scoring};
\draw[->,gray,dotted] (buyer.south) -- (nlpb.north);
\draw[->,gray,dotted] (engine.south) -- (nlpe.north);
\end{tikzpicture}
\caption{Agentic e-commerce as a micro-transaction market for \emph{verified information}. The buyer agent spends micro-payments (x402/AP2) to progressively unlock seller- and reviewer-supplied data; the heavy NLP lifting (dashed) is acquisition policy, pricing/negotiation, entity resolution, trust scoring, and grounding---not catalogue ranking.}
\label{fig:arch}
\end{figure*}
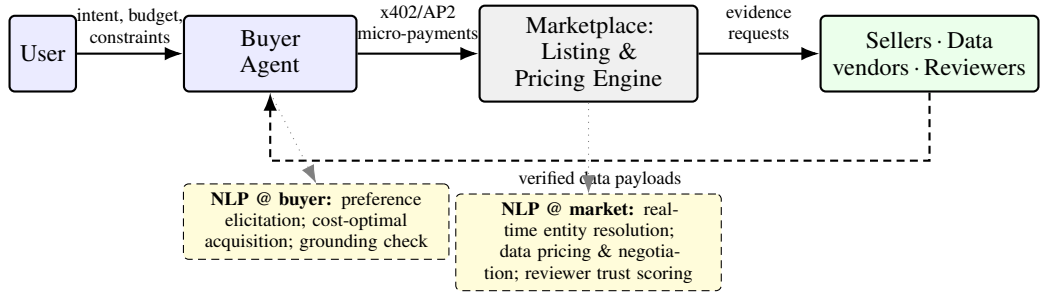

Figure~\ref{fig:arch} sketches the market. A user delegates intent, a budget, and constraints to a buyer agent; the buyer agent acquires evidence from a supply side of sellers, data vendors, and reviewers, settling micro-payments through a pricing engine over an x402/AP2 rail. We develop the mechanism below.

\subsection{Why not give the data away?}
Listing a product already costs money, and that fee funds the marketplace. We propose to make \emph{information itself} the unit of exchange. A listing exposes some attributes for free and prices the rest: the buyer agent knows the basics, asks for more, and pays per disclosure. Crucially, this is freemium \emph{with feedback}. A popular product---or one whose data reliably converts scrutiny into sales---can command higher prices for its information, and a seller-side pricing agent that learns to price each datum well captures proportionally more revenue. Demand for specific data is itself signal: an agent that pays to see torque figures reveals a different intent than one buying a service history, so the queries asked sharpen the seller's price prediction over time.

\subsection{What are the constraints, and how do agents negotiate them?}
Both sides operate under constraints. \emph{Hard} constraints are inviolable: a seller's reserve price or deadline to sell; a buyer's budget or a non-negotiable specification. \emph{Soft} constraints are tunable: how much to charge for each datum, how much to spend before escalating, how long to keep searching. Pricing and acquisition therefore become a negotiation \citep{liu2026agenticpay,zhang2026terms}: buyer and seller agents exchange offers over bundles of evidence, and a buyer rationally abandons a product the moment an unmet hard constraint is revealed---making \emph{what was asked, and when the buyer walked away} a rich pricing signal. A recurring design question is how often either agent should return to its human principal to relax a constraint.

\subsection{What is actually sold, and who verifies it?}
Beyond seller-attested attributes (materials, suppliers, audited sales and support metrics), the market admits third parties: independent test labs, and trusted reviewers who can be paid for an opinion or booked for a live session with the buyer. Because paid testimony invites fraud, reviewers carry a reputation/trust score settled by staking---borrowing the incentive logic of decentralised oracles and prediction markets \citep{peterson2020augur}---so that a reviewer later contradicted by outcomes loses standing and earning power.

\subsection{How should a buyer agent decide what to buy, and when to stop?}
On the buyer's side, every paid query is a tool call with a price tag, and the agent's competence is largely its \emph{acquisition policy}: which datum to buy next, and when to stop. This is the central open problem we return to in \S\ref{sec:challenges}; here we only note that it adds an explicit monetary objective to ordinary tool use---weighing the expected value of reduced uncertainty against both payment cost and the user's attention---and that the agent should surface the trade-off to its owner when the marginal cost of investigation approaches the value at stake.

\subsection{At the limit, why buy the finished product at all?}
Taken to its conclusion, an agent with enough verified information about a product's bill of materials and suppliers need not buy the finished good at all: it can source the components and commission assembly, buying the final product only when integration, warranty, or brand is worth the premium. This is information-driven competition at its sharpest---and a provocation about where the value, and the data, ultimately sit.

\section{Open Challenges for the Applied NLP Community}
\label{sec:challenges}

The vision is only as credible as the NLP it demands. We frame the open problems as five NLP tasks that recur as a purchase unfolds---acquire evidence, price and negotiate it, resolve and ground it, generate without fabricating, and model the buyer---each a place where the industry track is well placed to make progress.

\subsection{Cost-aware tool use}
Treating priced data and reviewer time as tools turns shopping into \emph{cost-aware tool use}: sequential decision-making under a budget that estimates the expected value of each available datum, buys the most cost-effective one, and stops when marginal value falls below price---or below the cost of bothering a human.\footnote{NLP work on curbing excessive, low-value tool calls is a natural starting point \citep{li2026entropy,schick2023toolformer}; our setting differs in making the objective explicitly monetary.} The hard part is calibration: an agent should pay for external data only out of \emph{epistemic necessity}---when its own parameters cannot resolve the uncertainty \citep{wang2025agent}---rather than reaching for a tool reflexively and buying facts it already reliably knows \citep{zeng2026tool}. And because each datum carries its own price, the familiar tool-use question of whether to call a tool is overtaken by a sharper one---\emph{which} tool to call when every option costs differently---so tool selection under heterogeneous prices becomes a far more pronounced problem here than in standard tool use, where external calls are free or uniformly cheap. Because the spend is committed before the answer is known, this also demands prospective control: triaging which queries to make under a fixed budget \citep{triage2026}, trading information gain against both payment cost and the user's attention.

\subsection{Negotiation dialogue}
Pricing a single datum, and negotiating bundles of evidence, casts product investigation as \emph{negotiation dialogue}---a long-standing NLP task \citep{lewis2017deal,he2018decoupling}---now at the intersection of dialogue and \emph{mechanism design} \citep{liu2026agenticpay,zhang2026terms}. It is also a \emph{game}: buyer and seller hold private valuations, each learns from \emph{which} data the other will buy or disclose, and both act strategically---so a competent agent reasons about best responses under incomplete information, not merely about a fair price. Progress is gated on \emph{evaluation}, since we cannot A/B test against real buyers cheaply or ethically. Existing benchmarks probe negotiation, economic decision-making, and agentic purchasing \citep{zhang2026terms,fish2026econevals,liu2026agenticpay,allouah2026agent,yu2026shopping}, yet none prices \emph{information itself} as a good bought under a micro-payment budget, nor scores the welfare and efficiency of paid-evidence acquisition. A practical path is to extend the community's Wizard-of-Oz and LLM-based user simulators \citep{budzianowski2018multiwoz,davidson2023usersim,sekulic2024usersim} into \emph{economic} simulators---populations of buyer and seller agents with private valuations and constraints---so that pricing and negotiation can be measured on welfare and calibration, not just deal rate.

\subsection{Entity resolution and knowledge-base construction}
For evidence to be priced and compared, it must be addressable and mean the same thing to both parties. Messy, multi-vendor evidence (``full service history, one owner from new'') must be resolved on the fly to a standardised product ontology, so that the same attribute from different sellers is comparable and individually priceable---entity and attribute resolution under latency and incomplete-schema pressure, which is what lets a buyer agent ask for, and pay for, exactly one missing field. The ontology and the rules for mapping onto it are market-specific---fitment rules for parts, failure modes for vehicles, compliance regimes for regulated goods---so an agent entering a particular market may dock a dedicated \emph{expert domain adapter} for resolution \citep{pfeiffer2021adapterfusion}, rather than rely on a single general resolver. Buyer and seller must also \emph{agree} they are discussing the same referent before a price means anything, a pragmatic-grounding step closer to consensus-and-repair than to one-shot record linkage. At marketplace scale this becomes continuous, provenance-tagged \emph{automated knowledge-base construction}: the lineage of never-ending extraction systems \citep{carlson2010nell}, but with every belief carrying a price and a verifiable source.

\subsection{Grounded generation and hallucination detection}
If a buyer agent fabricates an attribute---or a seller agent embellishes one---to keep a transaction alive, the market collapses: its entire premise is that \emph{paid} information is more trustworthy than free information. Generation must be constrained to a real, attestable inventory, extending grounding and citation methods for commercial dialogue \citep{zeng2025cite} toward verifiable provenance for every priced claim, with hallucination \emph{detection} a first-class check on both sides of a trade---and with persuasive optimisation \citep{salvi2026persuasion} bounded so that agents compete on evidence, not rhetoric.

\subsection{Preference elicitation and persona modelling}
Almost no agent today carries a model of the \emph{person}---their tastes, thresholds, and how and when they want to be notified---yet matching verified products to a buyer presupposes exactly this, making it the challenge that touches the user most directly.

\paragraph{Elicitation.} Durable preferences must be inferred from dialogue and behaviour \emph{without} interrogating the user: an agent that asks twenty questions before every purchase is worse than the storefront it replaces. The task is mixed-initiative preference elicitation that separates stable traits (brand affinities, a budget band, ethical constraints such as ``buy from the original manufacturer to support it'') from purchase-specific intent, and that knows when one clarifying question is worth more than another paid datum. Long-horizon, preference-grounded shopping benchmarks are a start \citep{yu2026shopping}, but most assume preferences are \emph{given} rather than discovered; the game-theoretic ``solicit-then-suggest'' model shows that multi-round solicitation reduces matching error far more than broadening final recommendations \citep{cao2026solicit}.

\paragraph{Representation and storage.} How the persona is stored is itself a research question. A natural-language or slot-structured profile is inspectable, editable, and portable, but leaks easily and must be reasoned over at every turn \citep{ramos2024transparent}; a parametric per-user ``personal model'' \citep{mindlab2026peft} is compact and expressive but opaque and hard to edit. One promising path is to use composable domain adapters \citep{pfeiffer2021adapterfusion} to dynamically dock category-specific expertise onto these personal weights. Preferences also drift, so any representation needs calibrated updating; recent work distils interaction history into compact preference memory for reranking \citep{peng2026memrerank}.

\paragraph{Alignment and notification timing.} The agent must align to what the user wants \emph{now}, not merely extrapolate past behaviour, and must resist a persona that adversarial listings have nudged. Alignment also governs \emph{when} to interrupt: today's assistants show an \emph{eagerness bias} that interrupts prematurely \citep{wang2026learning}, while the subjectivity of intervention timing means generalised triggers risk alarm fatigue \citep{modgil2026saturation}. This subsumes the earlier question of when to escalate to the principal to relax a constraint---a personalised judgement weighing the value of a clarification against its disruption cost.\footnote{Framing a contested purchase as a debate the user adjudicates \citep{irving2018debate,khan2024debating} is one way to keep a human in control of an agent acting on their behalf.} The same persona that personalises service becomes a liability if mis-stored, or if it leaks willingness-to-pay through role-coherence failures \citep{alavi2026role}---which we treat as an ethical question below.

\section{Conclusion}
\label{sec:conclusion}

We have argued that as e-commerce becomes agentic, the scarce and valuable good is not the ranking but \emph{verified information}, and that a micro-transaction market for that information---now technically feasible on agent-native payment rails---would reward genuine quality and sharpen competition. For applied NLP this is a redirection: away from optimising chat fluency and conversion, and toward the pipeline that turns dialogue into priced, grounded, negotiable evidence---cost-optimal acquisition, data pricing, real-time entity resolution, and hallucination-free value exchange. We invite the industry track to build and benchmark that pipeline.

\section*{Limitations}

This is a forward-looking position, not a deployed system or an empirical study; our claims about market efficiency and competition are hypotheses, not measurements. The vision presupposes adoption that may not materialise: sellers must be willing to expose verified data, buyers must trust agents with budgets, and agent-native payment rails (x402, AP2) are early and unevenly supported. Monetary figures in our running example (e.g., \pounds0.25 for a service-history digest) are illustrative only. Finally, the economic mechanisms we sketch---freemium data pricing and reviewer staking---are described at the level of a vision; their incentive compatibility and resistance to collusion require formal and empirical study that we do not provide here.

\section*{Ethical Considerations}

\paragraph{Consent and transparency.} A market that \emph{pays for} information must not become one that quietly \emph{extracts} it. Buyers should be told when their agent is purchasing data, from whom, and at what cost; sellers and reviewers should consent to what is exposed. We explicitly reject covert harvesting of a user's own data as a business model.

\paragraph{Privacy and PII.} Verified product data can carry personal data about people---a previous owner, a named reviewer, a support transcript. Any datum that is packaged and sold must be scrubbed of personally identifiable information unless its subject has consented; pipelines should treat PII removal as a first-class step, with attention to GDPR/CCPA obligations and to LLM memorisation of transaction data.

\paragraph{Manipulation and trust.} The channel that prices evidence can be attacked: SEO-style and prompt-injection attacks may hijack what an agent perceives, and persuasive optimisation \citep{salvi2026persuasion} may push agents to win on rhetoric rather than fact---hence our insistence on grounded value exchange and reputationally-staked reviewers \citep{peterson2020augur,irving2018debate}.

\paragraph{Persona storage and ownership.} Where the persona lives, and who controls it, is an ethical design decision, not an implementation detail. A profile rich enough to shop well is rich enough to exploit: a seller who can read it can price-discriminate up to the user's willingness to pay, and a leaked or stolen persona is a ready-made target for profiling and scams. Furthermore, agents are susceptible to ``role-coherence'' leakages, where natural-language persona descriptions or multi-turn conversational responses inadvertently expose a buyer's reservation prices and budget to strategic sellers \citep{alavi2026role}. We argue the persona should be \emph{user-held, portable, and inspectable}, with data minimisation and purpose limitation by default; and that the agent's own paid queries---which reveal the persona to sellers (\S\ref{sec:vision})---warrant privacy-preserving acquisition. Parametric personal models \citep{mindlab2026peft} also complicate the GDPR/CCPA right to erasure, since a preference encoded in weights cannot simply be deleted: reconciling personalised agents with rectification and the ``right to be forgotten'' is an open and pressing problem.

\paragraph{Power and access.} A data market can entrench incumbents who can afford to substantiate claims. Designing it so that small, honest sellers are \emph{rewarded} for transparency rather than priced out is a precondition for the ``truer competition'' we claim.

\bibliography{custom}

\end{document}